\documentclass[10pt,twocolumn,letterpaper]{article}

\usepackage{iccv}
\usepackage{times}
\usepackage{epsfig}
\usepackage{graphicx}
\usepackage{amsmath}
\usepackage{amssymb}
\usepackage{caption}
\usepackage{multicol}% http://ctan.org/pkg/multicols
\usepackage[linesnumbered,ruled]{algorithm2e}

\newcommand{\fii}[1]{\textit{FIVA}{#1}}

\newcounter{alphasect}
\def\alphainsection{0}

\let\oldsection=\section
\def\section{%
  \ifnum\alphainsection=1%
    \addtocounter{alphasect}{1}
  \fi%
\oldsection}%

\renewcommand\thesection{%
  \ifnum\alphainsection=1% 
    \Alph{alphasect}
  \else%
    \arabic{section}
  \fi%
}%

\newenvironment{alphasection}{%
  \ifnum\alphainsection=1%
    \errhelp={Let other blocks end at the beginning of the next block.}
    \errmessage{Nested Alpha section not allowed}
  \fi%
  \setcounter{alphasect}{0}
  \def\alphainsection{1}
}{%
  \setcounter{alphasect}{0}
  \def\alphainsection{0}
}%

\usepackage[breaklinks=true,bookmarks=false]{hyperref}

\iccvfinalcopy % *** Uncomment this line for the final submission

\def\iccvPaperID{****} % *** Enter the ICCV Paper ID here

\ificcvfinal\fi

\begin{document}

%%%%%%%%% TITLE
\title{FIVA: Facial Image and Video Anonymization and Anonymization Defense}

\author{Felix Rosberg$^{1, 2}$
\and
Eren Erdal Aksoy$^2$
\and
Cristofer Englund$^2$
\and
Fernando Alonso-Fernandez$^2$
\and
\small $^1$Berge Consulting, Gothenburg, Sweden
\and
\small $^2$Halmstad University, Halmstad, Sweden
\and
{\tt\small felix.rosberg@berge.io, \{eren.aksoy, cristofer.englund, 
 fernando.alonso-fernandez\}@hh.se}
}

\ificcvfinal\thispagestyle{empty}

\twocolumn[{%
\renewcommand\twocolumn[1][]{#1}%
\maketitle
\begin{center}
   \centering
    \captionsetup{type=figure}
    \includegraphics[width=0.99\linewidth]{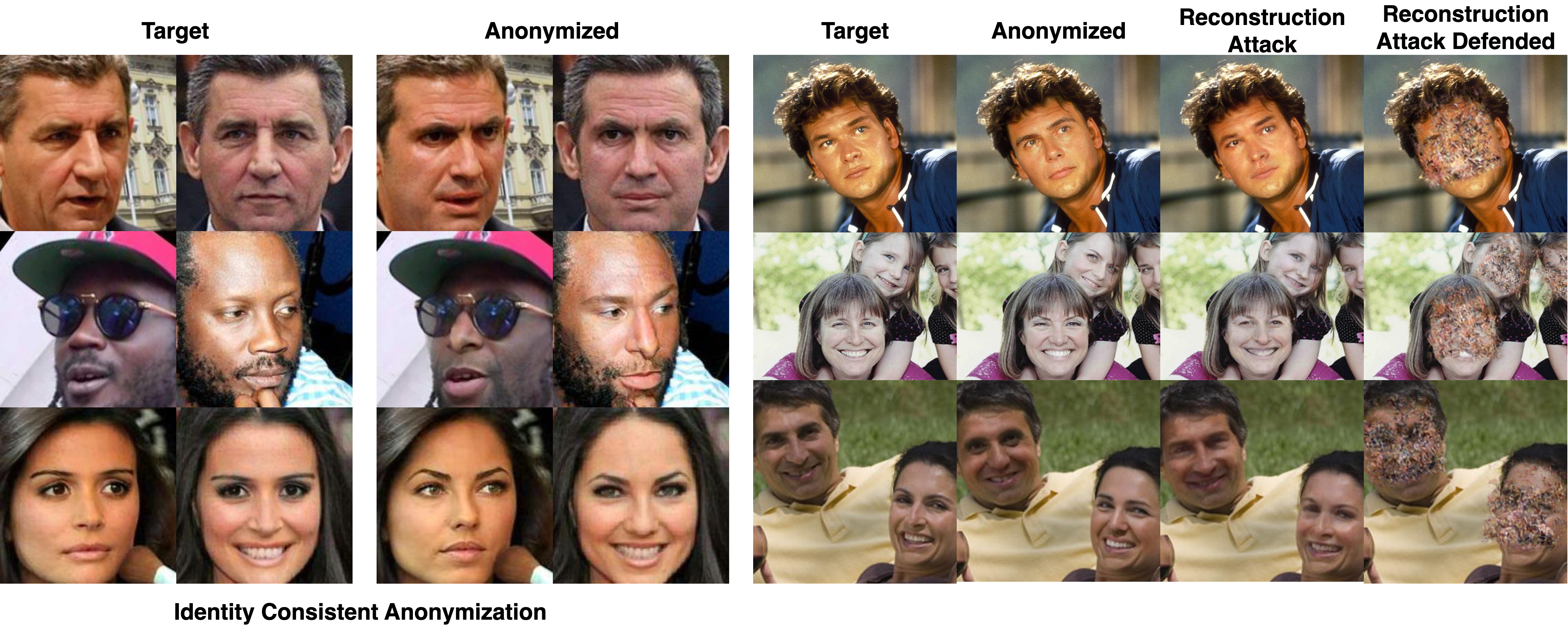}
    \captionof{figure}{Identity consistent anonymization generated by \fii~(left). Anonymization, reconstruction attacks and defense against reconstruction attacks (right). Reconstruction attacks allow for facial recognition to succeed as they manage to undo the anonymization. This is preventable by applying a small amount of noise to the anonymized image pixels, causing the reconstruction attack model to collapse.}
    \label{fig:best_results}
\end{center}%
}]

\thispagestyle{empty}

%%%%%%%%% ABSTRACT
\begin{abstract}
In this paper, we present a new approach for facial anonymization in images and videos, abbreviated as FIVA. 
Our proposed method is able to maintain the same face anonymization consistently over frames with our suggested identity-tracking and guarantees a strong difference from the original face.
FIVA allows for 0 true positives for a false acceptance rate of 0.001. 
Our work considers the important security issue of reconstruction attacks and investigates adversarial noise, uniform noise, and parameter noise to disrupt reconstruction attacks.
In this regard, we apply different defense and protection methods against these privacy threats to demonstrate the scalability of FIVA.
On top of this, we also show that reconstruction attack models can be used for detection of deep fakes. Last but not least, we provide experimental results showing how FIVA can even enable face swapping, which is purely trained on a single target image.
\end{abstract}

%%%%%%%%% BODY TEXT

\section{Introduction}

Privacy holds significant importance in various aspects of society, and data collection and storage are no exceptions. The increasing demand and interest in data, coupled with the implementation of recent regulations such as the General Data Protection Regulation, have made data anonymization a necessity. Numerous challenges arise where identity information becomes irrelevant, while attribute information remains crucial. Anonymization techniques aim to obscure, remove, or replace identity information with arbitrary pseudo-identities while preserving essential attribute information. However, obscuring or removing identity information through direct manipulation of the data distribution often results in the loss of significant attributes. For instance, techniques like blurring faces or replacing them with black boxes eliminate vital details such as eye gaze, pose, and expressions. In contrast, replacement-oriented methods focus on preserving key attributes while altering the identities of individuals.

In this study, we particularly concentrate on replacement-based approaches for anonymizing faces in both images and videos. 
We propose a novel anonymization method that utilizes target-oriented face-swapping models such as  FaceDancer \cite{FaceDancer} and SimSwap \cite{simswap} in conjunction with a model based on \cite{LIVEDEID}. 
It is here worth noting that our method can be implemented by using any existing target-oriented face manipulation model (e.g., \cite{simswap, faceshifter, FaceDancer, hififace}) that utilizes identity embeddings for facial manipulation. 
These models ensure strong consistency across video frames when the employed identity embedding is \emph{stable}. 
To both enable and enhance this consistency, we introduce a simple and efficient method that not only tracks identities but also samples fake identities. Thus, our proposed method enables robust anonymization and consistency across frames.

Furthermore, our research addresses an important aspect of security, specifically the vulnerability to reconstruction attacks, which has been under-investigated in facial anonymization. 
In a reconstruction attack, an adversarial model attempts to translate the anonymized face back to the original identity. We hypothesize and provide compelling evidence that anonymization models leave traces in the images that can be exploited for successful reconstruction attacks. To mitigate and verify this threat, we investigate the effectiveness of various noise types, including adversarial noise, uniform noise, and parameter noise, while disrupting the reconstruction attack. Additionally, we present results demonstrating how a reconstruction attack model can be utilized for deep fake detection.
 
Lastly, our contributions highlight that maximizing the identity distance is detrimental to privacy. This is because, state-of-the-art facial recognition models constrain embeddings to a hyper-sphere, which allows us to easily find the original identity by negating one of the embeddings.

\begin{figure*}[!t]
\centering
\includegraphics[width=0.85\textwidth]{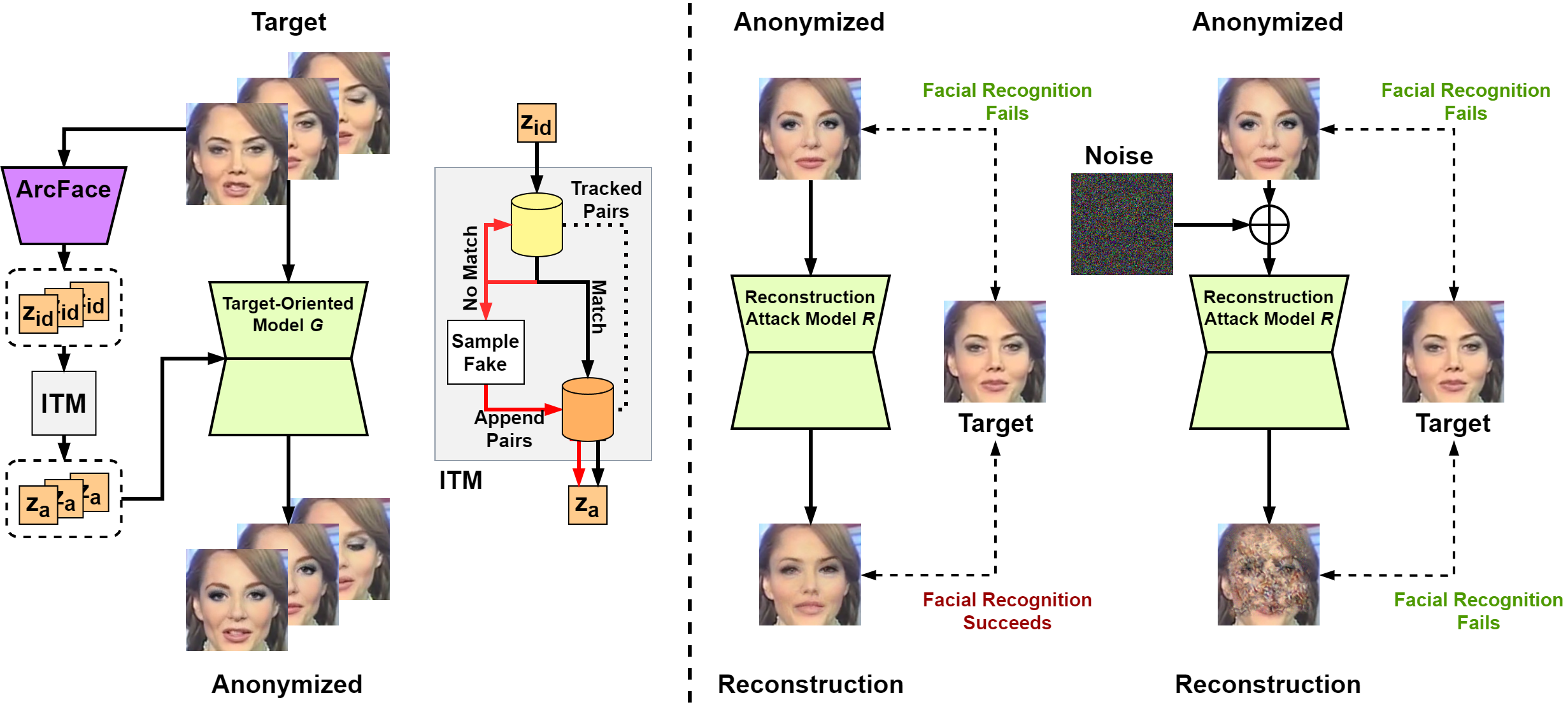}
\caption{Overview of our anonymizing pipeline: the Identity Tracking Module (ITM) (left)  and the implication of our reconstruction attacks (right). Here, $z_{id}$ and $z_{a}$ are the identity vector and the sampled fake identity vector, respectively. ITM checks if an identity exists and returns a corresponding fake identity (\emph{Match}). If not (\emph{No Match}), as indicated by the red arrows, we generate and/or save a fake identity and store the pair of vectors. We determine a match with the cosine distance and a manually chosen threshold (See Algorithm \ref{alg:itm}). The generator $G$, in this case, could be any target-orient facial manipulation model \cite{FaceDancer, simswap, hififace, facecontroller, faceshifter, DISENTANGLEDANON, LIVEDEID}. 
The reconstruction attack model $R$ shown on the right is learned in a black-box setting (No access to $G$). The model $R$  learns to reverse the transformation performed by $G$, thus, allowing for successful face recognition. By applying a small amount of noise to the anonymized image pixels, the reconstruction attacks can be defended, thus, yielding failed face recognition.
}
\label{fig:overview}
\end{figure*}

%-------------------------------------------------------------------------
\section{Related Work}

In this work, we focus on the direct manipulation of identity information within the data, such as identity masking and identity manipulation methods.
Manipulating identity involves masking or altering identity information to preserve privacy. In the context of image and video faces, two common masking approaches are blurring and obscuring with black boxes. While these methods may ensure strong privacy, they directly eliminate valuable information such as eye gaze, potentially affecting the data distribution. Naively training models on this distorted data may lead to the model becoming dependent on the introduced distortions. Therefore, an emerging approach is to leverage advancements in generative models to replace the identity with realistic faces \cite{MYFACEMYCHOICE, LIVEDEID, DEEPPRIVACY2, DISENTANGLEDANON, DEEPPRIVACY, ANONNET, CFANET, LEARNINGTOANON}. 
There exist various works, such as those by Ma \etal~\cite{CFANET}, Li \etal~\cite{DISENTANGLEDANON}, Li and Han \cite{ANONNET}, and Ren \etal~\cite{LEARNINGTOANON}, that directly employ face modification techniques. 
However, the current evaluations differ significantly, and none of these works focus on realism in a spatiotemporal context. Gafni \etal~\cite{LIVEDEID} utilize a face modification autoencoder network with a focus on spatiotemporal consistency. Their approach operates consistently across frames, generating a learned mask for occlusion awareness. However, there is no quantitative evaluation of the temporal consistency, and the network is trained to push the identity away while preserving attribute information, similar to the training approach of the face-swapping method FaceDancer \cite{FaceDancer}. DeepPrivacy and DeepPrivacy2 \cite{DEEPPRIVACY, DEEPPRIVACY2} employ a U-net-based model trained to inpaint a removed face, conditioned on pose information to preserve pose consistency. However, by completely removing the face, crucial information such as expression and eye gaze is lost. Temporal consistency is also disregarded, resulting in the generation of a new face for frames that differ slightly. \c{C}ift\c{c}i \etal~\cite{MYFACEMYCHOICE} utilize a face-swapping model, SimSwap \cite{simswap}, which we evaluate in this work using our proposed method. Therefore, our work complements theirs in this context.

Face-swapping techniques have emerged as a promising approach for facial anonymization in recent years, serving as the foundation for the direct manipulation of identity information. Specifically, a target-oriented face-swapping model is employed to facilitate the anonymization process.
face-swapping refers to the task of transferring a source face or identity onto a target face. Target-oriented methods, in particular, prove suitable for anonymization purposes as they typically employ identity embeddings from an identity encoder to directly manipulate the identity within the image \cite{simswap, faceshifter, FaceDancer, hififace}. Working with these identity embeddings is efficient and straightforward. Source-oriented approaches tend to struggle with lighting, textures and demands an actual fake image, which is costly \cite{3d_emily, fsgan, nirkin}. We demonstrate an approach that can be attached to an existing target-orient face-swapping model to allow for both image and video anonymization.

%-------------------------------------------------------------------------

\section{Method}
\label{s:method}

%%%%%%%%%%%%%%%%%%%%%%%%%%%%%%%%%%%%%%%%%%%%%%%%%%%%%%%%%%%%%%%%%%%%%%%%%%%%%%%%%%%%%%%%%%%%%%%%%%%%%%%%%%%%%%%%%%%%%%%%%%%%%%%%%%%%%%%%%%%%%%%%%%%%%%%%%%%%%%%%

\subsection{Network Architecture}
\label{ss:netarchitecture}
\fii~comprises a target-oriented encoder-decoder generator, a pre-trained ArcFace model \cite{arcface}, and the Identity Tracking Module (ITM) as illustrated in Figure \ref{fig:overview}. We here note that any target-oriented face manipulation model can be employed as the generator, for which ArcFace \cite{arcface} provides identity conditioning information. ITM is for tracking and sampling fake identities. For clarity and separation from other target-oriented models investigated, we will from here on denote the newly introduced model in this work as \fii.

\textbf{Generator (G):} Our generator model $G$ is based on the architecture presented in \cite{LIVEDEID}. We apply several important modifications to $G$, which are motivated by recent advancements in the literature and the need to address certain missing details in \cite{LIVEDEID}. First, the skip connection directly incorporates all feature maps from the encoder. Secondly, the identity information is broadcasted and concatenated with the bottleneck layer, preserving the spatial dimensions. The fully connected layers in the bottleneck are directly applied to the feature maps. Each encoder block utilizes a depth-wise convolution with a stride of 2 to downsample the feature maps, while the decoder block employs the pixel shuffle operation \cite{PIXELSHUFFLE} for upsampling. Similar to \cite{LIVEDEID}, our model generates an anonymized image, along with an automatically learned mask, which is designed to blend seamlessly with the target image (See supplementary material for a detailed overview).

\textbf{Discriminator:} Our discriminator, follows the design principles of discriminators used in FaceDancer \cite{FaceDancer}, HifiFace \cite{hififace}, and StarGAN-v2 \cite{stargan}. We employ the non-saturated GAN loss \cite{stylegan3, stylegan, stylegan2} for training.

\textbf{ArcFace:} In order to encourage the generator to deviate from the target image in terms of identity, we condition the bottleneck layer with identity vectors extracted from ArcFace \cite{arcface}, which utilizes a ResNet100 backbone \cite{resnet}.

\textbf{Reconstruction Attack Model (R):} Our reconstruction attack model, denoted as $R$, is a U-Net model. We choose to use similar encoder and decoder blocks to the ones described in \cite{LIVEDEID}. The main difference is that the attack model is not conditioned on any identity information and utilizes several skip connections. See supplementary materials for details.

\subsection{Identity Tracking Module (ITM)}
\label{ss:tracking}
When performing live anonymization in the wild, we need to keep track of the identities present in the video and the conditional information used to anonymize. To solve this, we simply use an Identity Tracking Module (ITM) that checks if the detected face has been observed before (Figure \ref{fig:overview}). If it has not, we sample a fake identity based on Equation \ref{eq:sample} as described in section~\ref{ss:samplefake}. If it has been observed, we skip the sampling and return the matching identity's corresponding fake identity. Let $z_{id}$ be the embedded identity vector, $\mathcal{T}_{id}$ is the stored previous identity vectors, $\mathcal{F}_{id}$ defines the dictionary containing the stored fake identity vectors $z_{a}$, $t$ denotes the threshold, $\textit{key\_pointer}$ represents the next key for adding new fake identities to $\mathcal{F}_{id}$, and $V_i$ is the sampling function. The process for tracking and generating new fake identities is then defined as shown in Algorithm \ref{alg:itm}. 

\begin{algorithm}
    \SetKwInOut{Input}{Input}
    \SetKwInOut{Output}{Output}

    \underline{ITM} $z_{id}$\;
    \Input{Extracted identity vector $z_{id}$}
    \Output{Fake identity $z_{a}$}
    $D$ = 1 - cosine\_similarity($z_{id}$, $\mathcal{T}_{id}$) \\
    $idx$ = argmin($D$) \\
    $d$ = $D$[$idx$] \\
    
    \eIf{$d < t$}
      {
        return $\mathcal{F}_{id}$[$idx$]\;
      }
      {
      $\mathcal{T}_{id}$.append($z_{id}$) \\
      $z_{a}$ = $V_i$($z_{id}$) \\
      $\mathcal{F}_{id}$[$\textit{key\_pointer}$] = $z_{a}$ \\
      $\textit{key\_pointer}$ = $\textit{key\_pointer}$ + 1 \\
        return $z_{a}$\;
      }
    \caption{Identity tracking algorithm for using consistent fake identities and generating new fake identities.}
    \label{alg:itm}
\end{algorithm}

\subsection{Sample Fake Identities}
\label{ss:samplefake}
For both our generator and target-oriented face-swapping models, we need to sample fake vectors, on which the models are conditioned. Considering those facial recognition models such as ArcFace \cite{arcface} and CosFace \cite{cosface} are trained to strengthen the cosine similarity between the same identity images, which is constraining the embeddings to a unit hypersphere, we can prove that the absolute opposite of a vector $z_{id}$ is simply put $-z_{id}$, see Equation \ref{eq:proof1} and \ref{eq:proof2}. This means that we can guarantee a strong anonymization (difference in identity) that facial recognition can not match correctly. However, this is a double-edged sword in which facial recognition guarantees a hit by searching for the most similar identity to $-z_{id}$. This means we need to sample identity information that is far away from \emph{both} $z_{id}$ and $-z_{id}$. This is done by preparing an anchor search space $\mathcal{S}_a$. In this regard, we extract average embeddings of identities in the VGGFace2 \cite{vggface2} train set. These are spherically interpolated with a shifted version of itself, thus creating an anchor search space of identity embeddings that all are a mix of two people (A visualization in a 3-dimensional coordinate system can be viewed in the supplementary material). This step can be repeated to a desired size of the anchor search space. We then search for an appropriate embedding as follows:

\begin{equation}
  z_a = \mathcal{S}_a[argmin(|cos(z_{id}, \mathcal{S}_a)| + m)]~,
  \label{eq:sample}
\end{equation}

where $z_{a}$ is a fake identity and $m$ is a margin to constrain the search result close to a specific distance. For instance, a margin $m$ of 0 would result in finding an embedding of a cosine similarity of 0.

Cosine similarity $cos(\theta)$ is defined as follows

\begin{equation}
  cos(\theta) = \frac{A \cdot B}{||A||||B||} ~,
  \label{eq:proof1}
\end{equation}

where $A$ and $B \in \mathcal{R}^D$, and the formula   is constrained between $-1$ and $1$. Assuming we measure the cosine similarity of $A$ with itself, we obtain

\begin{equation}
  cos(\theta) = \frac{A \cdot A}{||A||||A||} = 1~,
  \label{eq:proof2}
\end{equation}

which is equal to $1$, meaning exactly the same. Therefore, if we compare $A$ and $-A$ we obtain $cos(\theta) = -1$, meaning exactly opposite.

\subsection{Loss Functions}
The loss function has two components: one is for the face manipulation model and the other is for the reconstruction attack, described below.

\textbf{Face Manipulation Loss:} A combination of different losses is employed to train the generator model, such as  IFSR $\mathcal{L}_{ifsr}$ introduced in \cite{FaceDancer},  a cosine distance loss $\mathcal{L}_{a}$, L1 pixel-wise reconstruction loss, and L1 mask loss in \cite{LIVEDEID}. IFSR calculates the cosine distance between entire feature maps within an identity encoder to regularize the model to retent attribute information such as expression, pose, lighting and make-up. We modify IFSR slightly, instead of using a feature map for each residual block, we use a feature map for each resolution scale, and completely omit the proposed margins by setting them to 0. 
For identity manipulation, we choose to train it counterfactual, similar to \cite{LIVEDEID}. We use the cosine distance as:

\begin{equation}
    \mathcal{L}_{a} = 2 + cos(I(X_{t}), I(X_{a})) + cos(I(X_{t}), I(X_{\text{â}})) ~,
    \label{eq:antiidloss}
\end{equation}

where $cos(.)$ denotes the cosine similarity, $X_t$ is the target image, $X_a$ is the anonymized face, $X_\text{â}$ is the anonymized face blended with the target face and $I(.)$ is the pretrained ArcFace \cite{arcface}.

\textbf{Reconstruction Attack Loss:} To perform a reconstruction attack against a target-oriented face-swapping and/or anonymization model, we train a U-Net architecture to reconstruct the original image. The reconstruction attack model was trained using an L1 pixel-wise reconstruction loss, an identity loss, the same version of the IFSR loss $\mathcal{L}_{ifsr}$ from \cite{FaceDancer}, and an adversarial loss. The reconstruction loss is as follows:

\begin{equation}
  \mathcal{L}_{r} = ||X_t - R(X_c)|| ~,
\end{equation}

where $X_c$ is the anonymized/swapped target image, $X_t$ defines the unaltered target image of $X_c$, and $R(.)$ represents the reconstruction attack model. The identity loss is used to further enforce the original identity information:

\begin{equation}
  \mathcal{L}_{i} = 1 - cos(I(X_t), I(R(X_c))) ~,
  \label{eq:idloss}
\end{equation}

where $cos(.)$ denotes the cosine similarity. 
Once the reconstruction attack model is trained, we evaluate its capabilities in conjunction with the reconstruction attack vulnerability of the face-swapping model. In this case, we (1) sample a fake identity vector (see Section \ref{ss:samplefake}) for the attacked face-swapping model, (2) perform anonymization, (3) reconstruct the original identity with the reconstruction attack model, and finally (4) try to retrieve the original identity in the dataset using a separate identity embedding model, CosFace \cite{cosface}.

\subsection{Evaluation}
\label{ss:eval}

\textbf{Temporal Consistency:} Ideally, when performing anonymization on video, should the new identity be consistent over time. To gauge the identity consistency temporally, we first extract the face from $N$ frames from $M$ videos. Secondly, we extract the identity vector for each frame using CosFace \cite{cosface}. Finally, we calculate the pair-wise cosine distance between all frames and measure the mean standard deviation of these distances:

\begin{equation}
  \mathcal{M}^{\sigma}_{tc} = \frac{1}{M}\sum_{}^{M}\sqrt{\frac{1}{N^2}\sum_{i=1}^{N}\sum_{j=1}^{N}(D_{i,j} - \mu)^2} ~,
\end{equation}

where $D$ is the pair-wise cosine distance matrix and $\mu$ is the mean distance of $D$. We also report the mean of the mean distances in $D$, denoted as $\mathcal{M}^{\mu}_{tc}$. This lets us represent the mean variation of the anonymized face when the identity is the same as the input for the anonymization algorithm.

\textbf{Anonymization:} We follow previous works \cite{LIVEDEID, CIAGAN, DISENTANGLEDANON, towards_privacy_fastzero} for evaluating how well our method manages to hide the identity of the target in contrast to other approaches.
In most of the face-swapping algorithms \cite{simswap, faceshifter, FaceDancer, hififace, facecontroller} and reconstruction attack approaches, the identity performance is evaluated using the percentage of successful retrieval of the original source identity. 
We here rather report the percentage of successful retrieval of the target identity, where the goal is to negate the recapturing of the identity. 
Another difference is that successful retrieval of the target identity is not necessarily successful in a practical setting. 
More specifically, we follow common practices from recognition tasks and reject the retrieved identity as a success if and only if the closest identity is the target identity with a distance that is still larger than a practical threshold. In this case, the threshold is 0.63 for a false acceptance rate of 0.001.

\textbf{Adversarial Defense:} The theory for why reconstruction works is because it learns a function in the image that the anonymization model ``\textit{applies}". Therefore, we investigate potential approaches that could disrupt that function. For defending against reconstruction attacks we investigate adversarial attacks, more specifically, the fast gradient sign attack \cite{FSGM}, against the reconstruction attack model, standard uniform noise, and Gaussian noise applied to the parameters of the model.

\subsection{Reconstruction Attacks as Deep Fake Detectors}
Since the reconstruction attack model is trained to take a manipulated face image as input and perform an alteration in the image itself to restore the original identity, we investigate the cases when the input is not manipulated by a target-oriented face-swapping model.
Since our hypothesis states that the reconstruction attack model learns a function in the image that the anonymization model or face-swapping model leaves behind,  we expect that images that have not been manipulated cannot be changed by the reconstruction attack model. This claim is supported by the experimental results shown in Section \ref{s:results}. 
Because of this behavior, we can use the reconstruction attack model as a deep fake detector. This can be easily done by measuring the cosine distance between identity embeddings of the input image and output image, similar to Eq. \ref{eq:idloss}.
A high distance value would indicate a deep fake, while a low distance does not.

\section{Results}
\label{s:results}
\textbf{Implementation Details:} \fii~is trained on the dataset VGGFace2~\cite{vggface2}. All faces are aligned with five-point landmarks extracted with RetinaFace~\cite{retinaface}. The alignment is performed to match the input into ArcFace~\cite{arcface}. We used the Adam~\cite{adam} optimizer with $\beta_1 = 0.5$, $\beta_2 = 0.99$, a learning rate of 0.0001, and exponential learning rate decay of $0.97$ every 100K steps. 
The target ($X_t$) face is distorted during training with random rotation of 10 degrees and small random zooms are applied before being fed to \fii. The input $X_t$ of ArcFace remains undistorted. Image resolution is $256\times256$.
%%%%%%%%%%%%%%%%%%%%%%%%%%%%%%%%%%%%%%%%%%%%%%%%%%%%%%%%%%%%%%%%%%%%%%%%%%%%%%%%%%%%%%%%%%%%%%%%%%%%%%%%%%%%%%%%%%%%%%%%%%%%%%%%%%%%%%%%%%%%%%%%%%%%%%%%%%%%%%%%

\subsection{Quantitative Results}
\label{s:quant}
Exhaustive experiments are conducted to demonstrate the effectiveness of our proposed \fii~model together with target-oriented face-swapping models.
Table \ref{t:compare} contains quantitative results on the dataset FaceForensic++ \cite{faceforensics++}. This table not only compares \fii~with previous works, but also acts as an ablative benchmark for highlighting the effectiveness of our proposed anchor sampling and ITM modules. 
The metrics evaluated are identity retrieval (ID), negated identity retrieval ($\neg$ID), reconstruction attack identity retrieval (RA), and temporal consistency (See Section \ref{ss:eval}).
As shown in Table \ref{t:compare}, both \fii, SimSwap~\cite{simswap}, and FaceDancer~\cite{FaceDancer} achive the lowest ID scores and thus allow for strong anonymity of faces.

%======                                                                     =======
%======    Main experiment notes,                                           ======= 
%======    currently baking the rec attack results in her                   =======
\begin{table}[t!]
\caption{Quantitative experiments on FaceForensics++~\cite{faceforensics++}. Evaluated with identity retrieval (ID), negated identity retrieval ($\neg$ID, searching for a match with $-z_{id}$), and reconstruction attack (RA) identity retrieval. Temporal identity consistency $\mathcal{M}_{tc}$ calculated using 10 frames per video. The divide in the table separates inpainting-based methods from target-oriented ones. The $\times$ indicates that RA is not applicable to the corresponding method. +Sampling means we used the anchor sampling method to assign anonymized identities (See Equation \ref{eq:sample}), while +ITM indicates both the anchor sampling and tracking (See Algorithm \ref{alg:itm} and Figure \ref{fig:overview}). The $\downarrow$ indicates lower is better.}
\label{t:compare}
\begin{center}
\resizebox{0.49\textwidth}{!}{
\begin{tabular}{p{4.3cm}ccccc}
\hline
Method & ID$\downarrow$ & $\neg$ID$\downarrow$  & RA$\downarrow$ & $\mathcal{M}^{\mu}_{tc}\downarrow$  & $\mathcal{M}^{\sigma}_{tc}\downarrow$\\
\hline
Real Data & - & - & - & 0.150 & 0.074\\
\hline

CIAGAN~\cite{CIAGAN} & 0.035 & \textbf{0.000} & $\times$ & 0.521 & 0.220\\
CIAGAN~\cite{CIAGAN} + ITM & 0.030 & \textbf{0.000} & $\times$ & 0.300 & 0.151\\

DeepPrivacy~\cite{DEEPPRIVACY} & 0.004 & \textbf{0.000} & $\times$ & 0.359 & 0.184\\

\hline

CFA-Net~\cite{CFANET} & 0.012 & N/A & N/A & N/A & N/A\\

SimSwap~\cite{simswap} + Sampling & 0.002 & \textbf{0.000} & \textbf{0.994} & 0.607 & 0.345\\
SimSwap~\cite{simswap} + ITM & 0.002 & \textbf{0.000} & \textbf{0.994} & 0.084 & 0.051\\

FaceDancer~\cite{FaceDancer} + Sampling & \textbf{0.000} & \textbf{0.000} & 0.999 & 0.556 & 0.314\\
FaceDancer~\cite{FaceDancer} + ITM & \textbf{0.000} & \textbf{0.000} & 0.999 & 0.186 & 0.141\\

FIVA & \textbf{0.000} & 0.966 & 0.998 & 0.227 & 0.101\\
FIVA + Sampling & \textbf{0.000} & \textbf{0.000} & 0.996 & 0.550 & 0.310\\
FIVA + ITM & \textbf{0.000} & \textbf{0.000} & 0.996 & \textbf{0.075} & \textbf{0.041}\\

\hline
\end{tabular}
}
\end{center}
\end{table}

Our proposed method of sampling fake identities shows that we can avoid identity leakage by searching for $-z_{id}$.
Recalling that \fii~was trained counterfactually and can, in theory, anonymize without any sampling, Table \ref{t:compare} shows a successful identity retrieval rate ($-z_{id}$) of 96.6\%  ($\neg$ID) even if anchor sampling is \emph{not} employed. 
Note that SimSwap and FaceDancer are primarily designed as face-swapping approaches and both require an additional sampling process.
Our proposed ITM approach allows for strong temporal consistency in all the target-oriented methods, as shown in Table \ref{t:compare} where the $\mathcal{M}_{tc}$ scores drop once ITM is added to SimSwap~\cite{simswap},  FaceDancer~\cite{FaceDancer}, and \fii.

%======                                                                     =======
%======    Main experiment notes,                                           ======= 
%======    currently baking the rec attack results in her                   =======
\begin{table}[t!]
\caption{Quantitative identity retrieval experiments on LFW~\cite{LFW}. CFA-Net \cite{CFANET} and Gafni et al. \cite{LIVEDEID} demonstrate the true positive rate for a false acceptance rate of 0.001 using FaceNet \cite{FACENET} as the facial recognition model. We evaluate the remaining methods with CosFace and a threshold of 0.63 (Cosine \emph{distance}), for a false acceptance rate of 0.001. The $\downarrow$ indicates lower is better.}
\label{t:compare2}
\begin{center}
\begin{tabular}{p{3.5cm}cc}
\hline
Method & ID$\downarrow$\\
\hline
Gafni et al. \cite{LIVEDEID} & 0.035\\
CIAGAN~\cite{CIAGAN} & 0.034\\
CFA-Net~\cite{CFANET} & 0.012\\
DeepPrivacy~\cite{DEEPPRIVACY} & 0.002\\
FaceDancer~\cite{FaceDancer} + ITM & 0.002\\
SimSwap~\cite{simswap} + ITM & 0.001\\
FIVA + ITM & \textbf{0.000}\\
\hline
\end{tabular}
\end{center}
\end{table}

Next, we compare the performance of \fii, SimSwap~\cite{simswap}, and FaceDancer~\cite{FaceDancer} as anonymizer (utilizing ITM for tracking and anchor sampling) with previous work of the LFW benchmark \cite{LFW} in Table \ref{t:compare2}. 
The used facial recognition model differs here, however, we claim that the comparison is still valid as we use a more powerful model, CosFace \cite{cosface}, for identity retrieval. Table \ref{t:compare2} further demonstrates the effectiveness of both \fii~and ITM in providing robust anonymization.

%======                                                                     =======
%======    Investigating Reconstruction attack                                         ======= 
%======                                                                     =======
\begin{table}[b!]
\caption{Defense against reconstruction attack in \fii~, evaluated on FaceForensics++~\cite{faceforensics++}. Adversarial Defense in the form of a fast sign gradient method. Noise Defense just adds regular uniform noise to the image. Parameter Noise means adding a small Gaussian noise to the parameters. We report the fraction of successful retrievals of the original identity after applying the reconstruction attack. $\epsilon$ highlights how much the noise was scaled. The $\downarrow$ indicates lower is better. Black-box means it does not need access to the reconstruction attack model.}
\label{t:defense}
\begin{center}
\begin{tabular}{p{3.0cm}ccc}
\hline
Method & $\epsilon$ & ID$\downarrow$ & Black-box\\
\hline
Parameter Noise & 0.10 & 0.442 & \textbf{yes}\\
Adversarial Defense & 0.15 & \textbf{0.002} & no\\
Noise Defense & 0.15 & 0.004 & \textbf{yes}\\
\hline
\end{tabular}
\end{center}
\end{table}

Finally, we claim that  \fii~and other target-orient approaches leave a trace in the image that allows for an adversarial network to learn to reconstruct the original identity. This is highlighted by the lower scores in column 3 in Table \ref{t:compare}. 
To strengthen this claim and demonstrate potential approaches, we show in Table \ref{t:defense}  that a disrupting noise can collapse the output of the reconstruction attack model. 
Furthermore, in Figure \ref{fig:attack_success} we investigate how many pixels are needed to disrupt the reconstruction attack model. 
As shown in Figure \ref{fig:attack_success}, for \fii, 47\% noised pixels yielded the best disruption of the reconstruction attack model, but as little as 10\% causes a severe collapse of the reconstruction attack model.
Figure \ref{fig:defense} provides qualitative results of the reconstruction attack, different defenses, and anonymization using \fii.

\begin{figure}[!t]
\centering
\includegraphics[width=0.5\textwidth]{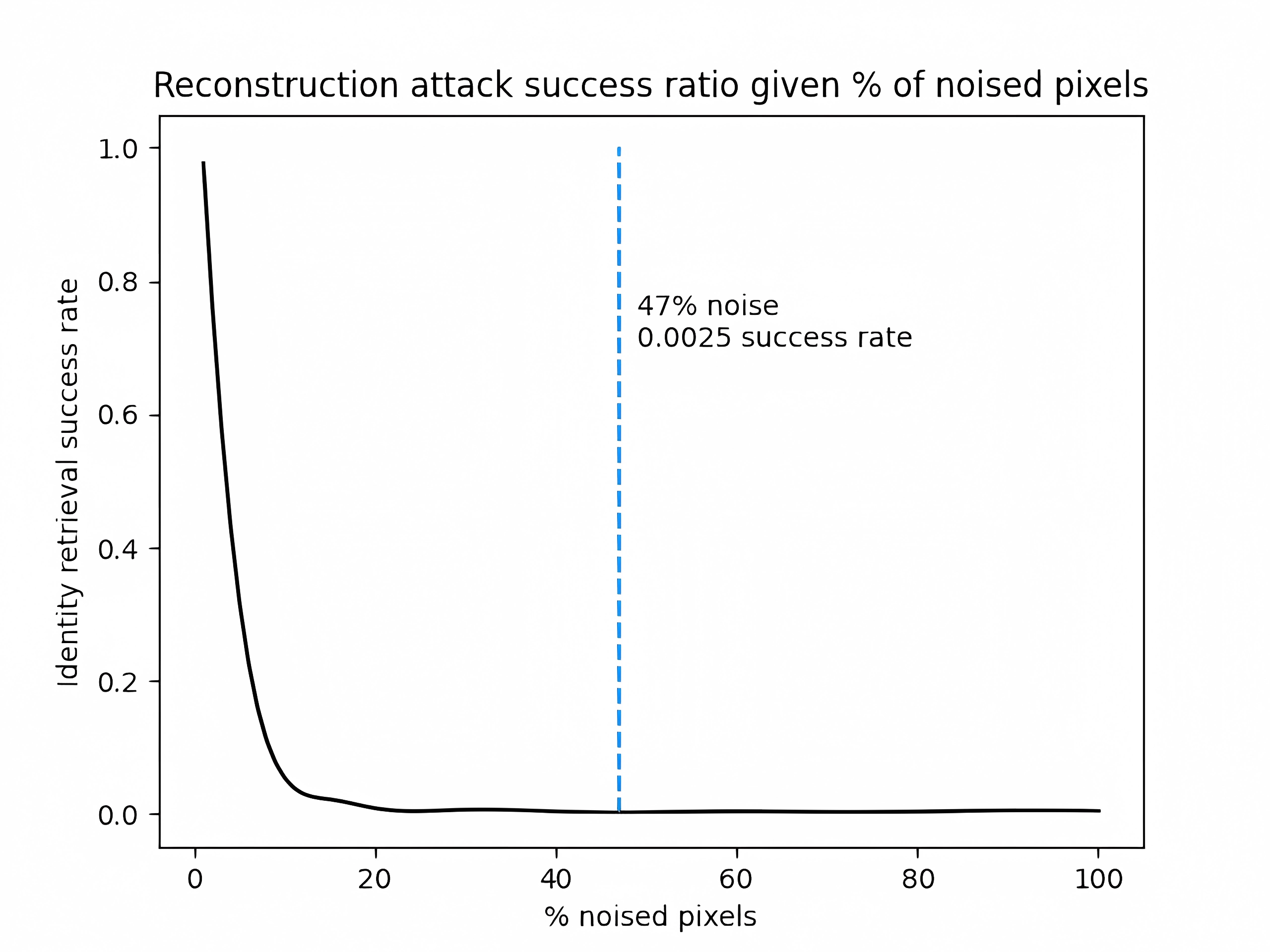}
\caption{
Identity retrieval success rate of the reconstruction attack, depending on the number of noisy pixels. This graph demonstrates the identity retrieval success rate for \fii~while using the uniform noise. The blue dashed line points out the \% which prevents reconstruction attacks best (0.0025).
}
\label{fig:attack_success}
\end{figure}

\begin{figure}[!b]
\centering
\includegraphics[width=0.47\textwidth]{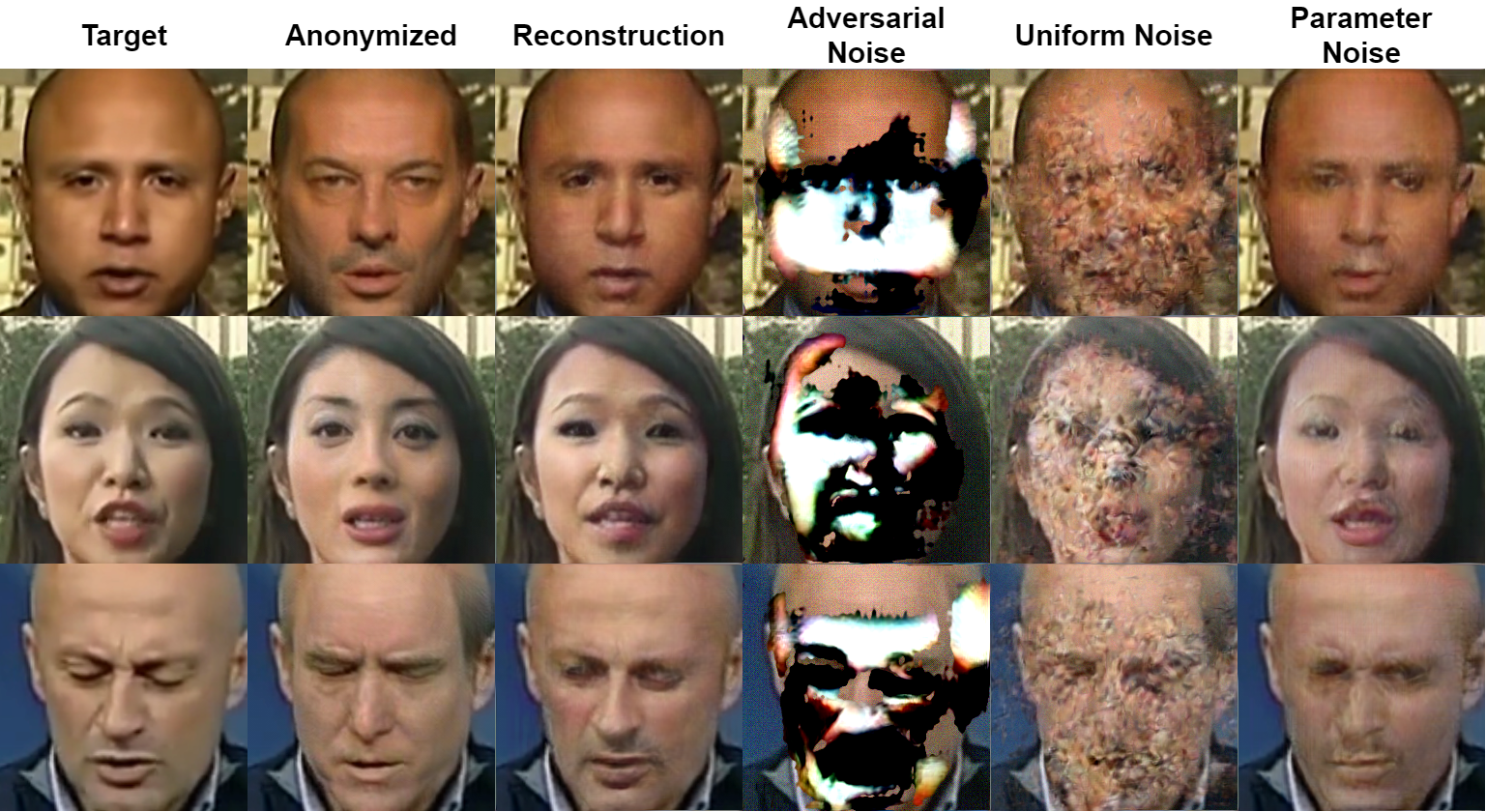}
\caption{Qualitative results of reconstruction attack, different defenses and anonymization using \fii.}
\label{fig:defense}
\end{figure}

%%%%%%%%%%%%%%%%%%%%%%%%%%%%%%%%%%%%%%%%%%%%%%%%%%%%%%%%%%%%%%%%%%%%%%%%%%%%%%%%%%%%%%%%%%%%%%%%%%%%%%%%%%%%%%%%%%%%%%%%%%%%%%%%%%%%%%%%%%%%%%%%%%%%%%%%%%%%%%%%

\subsection{Reconstruction Attack as Deepfake Detection}
The success of reconstructing the original identity from target-oriented face-swapping or anonymization approaches raises the question of what happens if one inputs an image that is not manipulated. We notice that the reconstruction attack model does not really change anything at all with these images. As demonstrated in Figure \ref{fig:dfd}, a potential approach for deep fake detection is simply measuring the cosine distance of the CosFace embeddings between the input image $X$ and the output image $'X$. We note that a threshold of 0.6 would reach a false positive of almost 0. One drawback is that the reconstruction attack model is as of now not model agnostic.

\begin{figure}[!ht]
\centering
\includegraphics[width=0.5\textwidth]{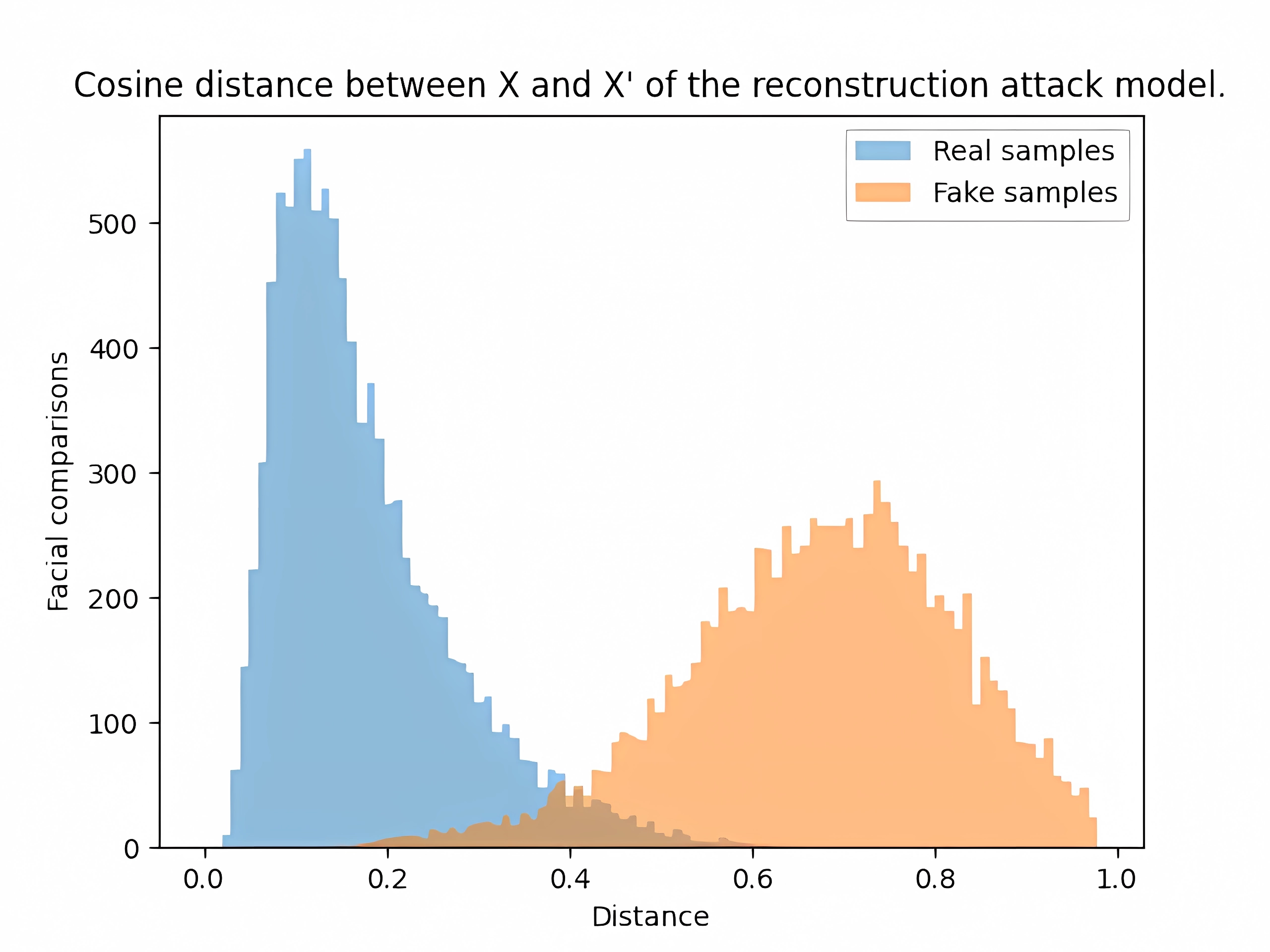}
\caption{Cosine distance distributions between the input ($X$) and output ($'X$) identity vectors of the reconstruction attack for genuine images (Real samples) and manipulated images (Fake samples).}
\label{fig:dfd}
\end{figure}

%======                                                                     =======
%======    Zero-shot Face Swaps                                        ======= 
%======                                                                     =======
\begin{table}[b!]
\caption{Quantitative face-swapping comparisons on FaceForensics++~\cite{faceforensics++}. The $\downarrow$ indicates lower is better, while $\uparrow$ indicates higher is better.}
\label{t:fs}
\begin{center}
\begin{tabular}{p{3.1cm}cc}
\hline
Method & ID$\uparrow$ & Pose$\downarrow$\\
\hline
FaceSwap~\cite{faceswap} & 54.19 & 2.51\\
FaceShifter~\cite{faceshifter} & 97.38 & 2.96\\
MegaFS~\cite{megaface} & 90.83 & 2.64\\
FaceController~\cite{facecontroller} & 98.27 & 2.65\\
HifiFace~\cite{hififace} & 98.48 & 2.63\\
SimSwap~\cite{simswap} & 92.83 & \textbf{1.53}\\
FaceDancer~\cite{FaceDancer} & 98.84 & 2.04\\
FIVA (Ours) & \textbf{99.25} & 2.16\\
\hline
\end{tabular}
\end{center}
\end{table}

%%%%%%%%%%%%%%%%%%%%%%%%%%%%%%%%%%%%%%%%%%%%%%%%%%%%%%%%%%%%%%%%%%%%%%%%%%%%%%%%%%%%%%%%%%%%%%%%%%%%%%%%%%%%%%%%%%%%%%%%%%%%%%%%%%%%%%%%%%%%%%%%%%%%%%%%%%%%%%%%
\subsection{Face-swapping using FIVA}
Since \fii~is trained to drive the identity away using the cosine distance (Eq. \ref{eq:antiidloss}), we hypothesize that \fii~can also perform face-swapping even if it is not the main goal. 
Table \ref{t:fs} shows quantitative comparisons with the state-of-the-art face-swapping methods on the dataset FaceForensics++~\cite{faceforensics++}. We follow the same evaluation protocols in \cite{simswap, faceshifter, FaceDancer, hififace} and obtain that \fii~reaches state-of-the-art performance for the identity transfer (ID). 
This concludes that target-oriented face-swapping methods can actually be trained in a counterfactual way, which eliminates the need for sampling pairs of faces during training. 
However, we  note that \fii~naturally tends to keep attributes such as gender, ethnicity, and face shape as  shown in Figure \ref{fig:faceswap}. 
Consequently, the obtained swapped faces are not perceptually convincing for humans but rather are for facial recognition models. 
This feature is useful for the task of anonymization, and further addresses ethical questions in regard to deep fakes in the context of anonymization. 
We elaborate more on this matter in the supplementary materials.

\begin{figure}[!t]
\centering
\includegraphics[width=0.3\textwidth]{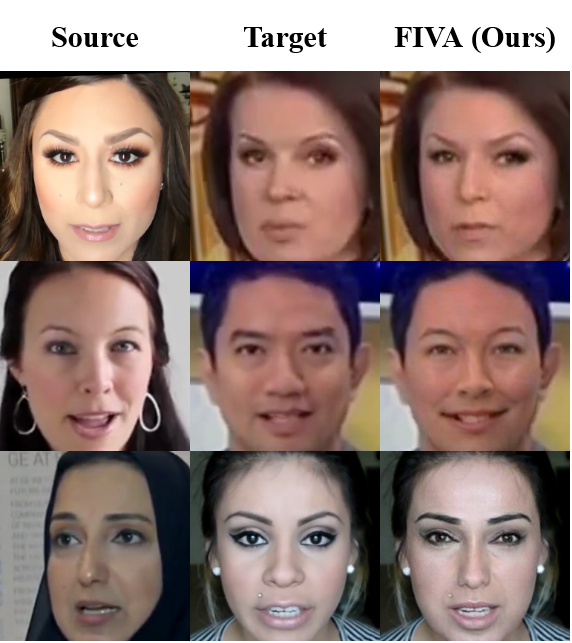}
\caption{Face swapping results using \fii.}
\label{fig:faceswap}
\end{figure}

%%%%%%%%%%%%%%%%%%%%%%%%%%%%%%%%%%%%%%%%%%%%%%%%%%%%%%%%%%%%%%%%%%%%%%%%%%%%%%%%%%%%%%%%%%%%%%%%%%%%%%%%%%%%%%%%%%%%%%%%%%%%%%%%%%%%%%%%%%%%%%%%%%%%%%%%%%%%%%%%
\subsection{Qualitative Results}
\label{s:qual}

\begin{figure}[!b]
\centering
\includegraphics[width=0.45\textwidth]{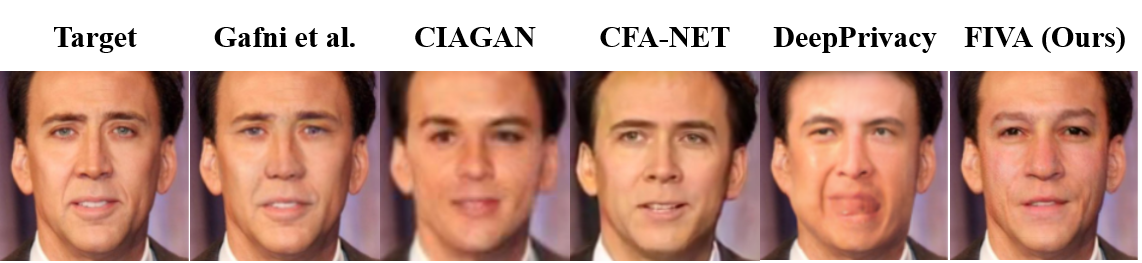}
\caption{Qualitative comparison between Gafni et al. \cite{LIVEDEID}, CIAGAN \cite{CIAGAN}, CFA-NET \cite{CFANET}, DeepPrivacy \cite{DEEPPRIVACY}, and \fii.}
\label{fig:qual_comp}
\end{figure}

For qualitative evaluation, we compare the output of \fii~with previous works, as shown in Figures \ref{fig:qual_comp} and \ref{fig:qual_comp_temporal}. 
In-depth comparisons are performed with the already available models such as Gafni et al. \cite{LIVEDEID}, CIAGAN \cite{CIAGAN}, CFA-NET \cite{CFANET}, and DeepPrivacy \cite{DEEPPRIVACY}. 
Images in Figures \ref{fig:qual_comp} and  \ref{fig:qual_comp_temporal} clearly show that CFA-NET struggles with maintaining the color and eye-gaze, whereas CIAGAN has issues with the resolution of the image and returns rather low-quality outputs. 
DeepPrivacy often produces artifacts while struggling with eye-gaze and facial expression. 
Gafni et al.~\cite{LIVEDEID} together with CFA-NET and \fii,~is the only approach that demonstrates successful results on video. 
In this particular frame, the change is arguably small. CFA-NET does use similar identity control as target-oriented face-swapping and \fii, which means that they need to manually assign identity embedding for each face in a video, restricting their use for in-the-wild anonymization. 
This problem is solved by our contribution ITM, whcih can be directly applied to their work. 
Gafni et al.~\cite{LIVEDEID} train their model in a counterfactual fashion, and produce stable videos without the need to track identities. 
This is true for \fii~as well, however, as pointed out in section \ref{ss:samplefake}, a maximized cosine distance allows facial recognition to find the identity by searching for $-z_{id}$. 
We visualize and discuss this further in the supplementary material. For video results, we refer to the supplementary material.

\begin{figure}[!ht]
\centering
\includegraphics[width=0.4\textwidth]{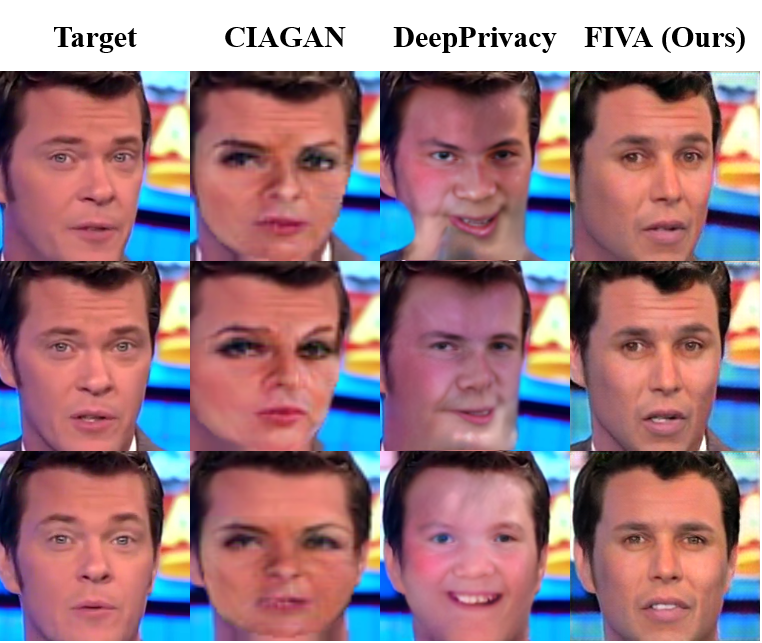}
\caption{Qualitative temporal comparison between CIAGAN \cite{CIAGAN}, DeepPrivacy \cite{DEEPPRIVACY} and \fii. Note we used ITM for tracking the identity in CIAGAN. Video results demonstrating this can be found in the supplementary material.}
\label{fig:qual_comp_temporal}
\end{figure}

%%%%%%%%%%%%%%%%%%%%%%%%%%%%%%%%%%%%%%%%%%%%%%%%%%%%%%%%%%%%%%%%%%%%%%%%%%%%%%%%%%%%%%%%%%%%%%%%%%%%%%%%%%%%%%%%%%%%%%%%%%%%%%%%%%%%%%%%%%%%%%%%%%%%%%%%%%%%%%%%

\section{Conclusion}
In this work, we introduce a new facial anonymization framework \fii, which together with our proposed identity sampling and identity tracking, reaches state-of-art performance for facial anonymization for both video and images.
We also show that target-oriented models are very easy to attack and, thus, introduce adversarial models that can completely undo the masked identity in the frame. 
To the best of our knowledge, this potential security issue has so far not been addressed. 
We furthermore demonstrate that regardless of what the attack model looks for in the image, it can be disrupted by noise. 
We expect this to become a cat-and-mouse game where attack models can learn to ignore the noise, thus making anonymization more challenging.
The attack model can also be used as deep fake detector, since the model does not change anything in the image when the input has not been manipulated.
Last but not least, \fii~is also capable of face-swapping, reaching state-of-the-art performance for identity transfer, thus demonstrating its excellent control over identity information. Interestingly it does so, while keeping the changes to a minimum.

{\small
\bibliographystyle{ieee_fullname}
\bibliography{egbib}
}

\newpage

\onecolumn
\section*{Supplementary Material}
In this section we show further details, including fake identity sampling and preservation of gender and ethnicity attributes. You can also find video results here\footnote{\url{https://drive.google.com/drive/folders/1RY3ozfdwP5v13cTELRJe2gEdXrrfaUMP}}. You will also find detailed network structures of both the reconstruction attack model and the FIVA generator.

\begin{center}
   \centering
    \captionsetup{type=figure}
    \includegraphics[width=1.0\linewidth]{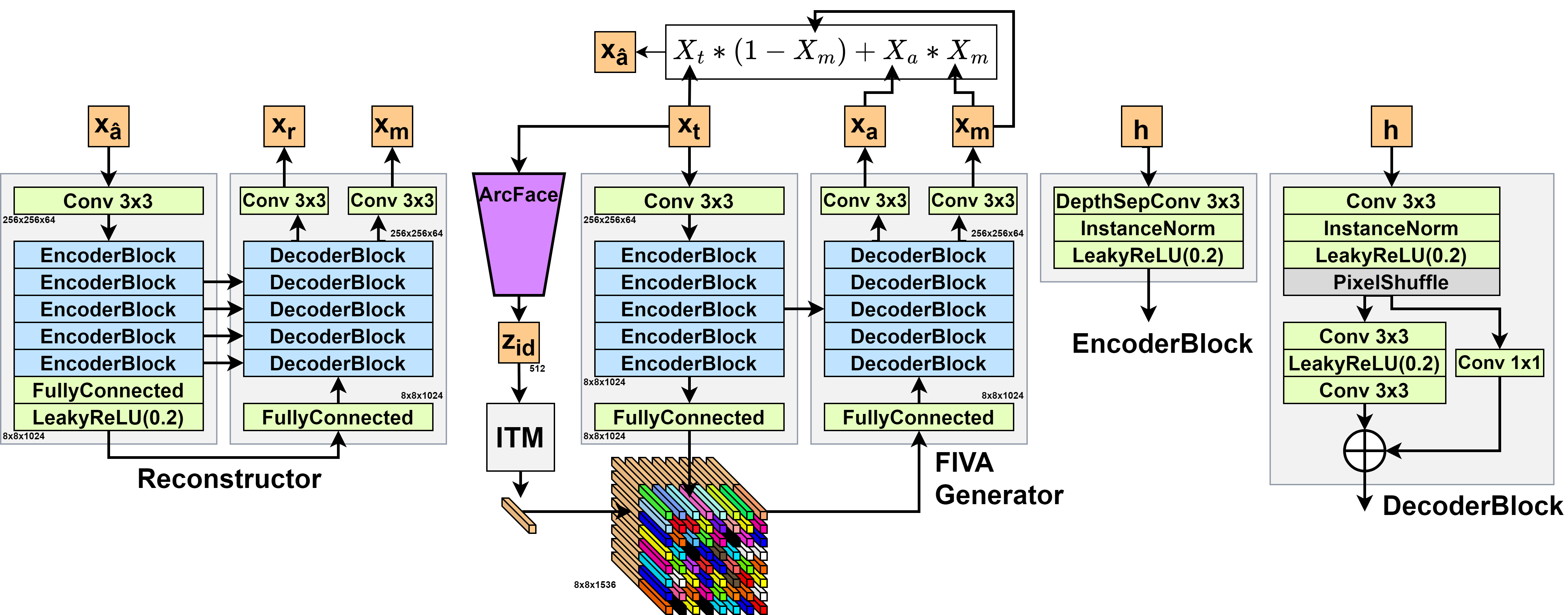}
    \captionof{figure}{Overview of the reconstruction attack model and the \fii~generator. The skip connections are concatenated with the upsampled feature maps from the previous DecoderBlock before being passed into the next DecoderBlock.}
    \label{fig:overview}
\end{center}%

\begin{alphasection}
\section{Fake Identity Sampling Details}

\begin{figure}[ht!]
\centering
\includegraphics[width=0.8\textwidth]{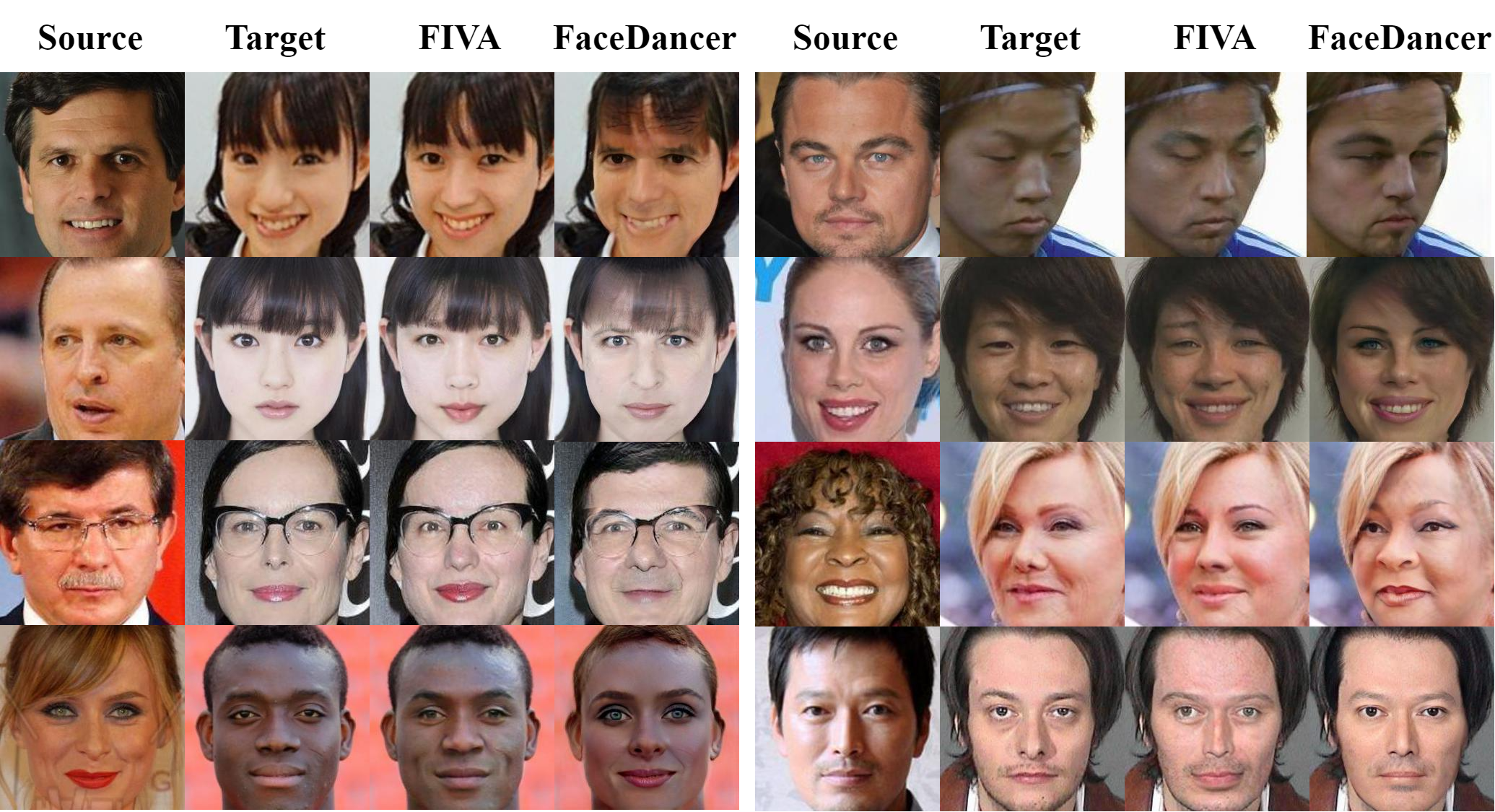}
\caption{Qualitative comparison between \fii~and FaceDancer for gender and ethnicity retention.}
\label{fig:attri}
\end{figure}

\begin{figure}[ht!]
\centering
\includegraphics[width=0.4\textwidth]{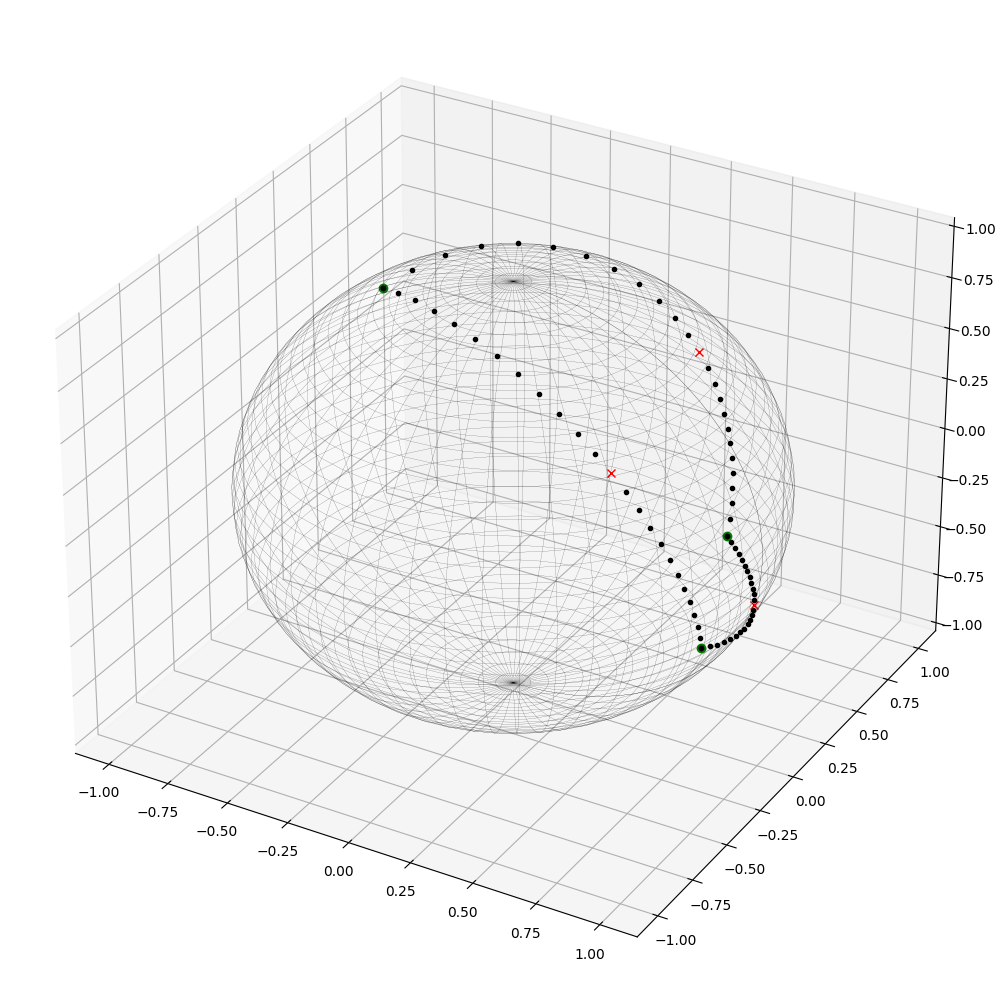}
\caption{Illustration of how we create the anchor vectors that we sample fake identities from. The green points illustrates an identity extracted by ArcFace, the black dots shows the spherical interpolation path on the unit sphere and the red crosses represent the resulting vectors used in the anchor set which becomes a mix of actual identities.}
\label{fig:slerp}
\end{figure}

In Figure \ref{fig:slerp} we visualize in a 3d projection how the anchor set of mixed identities is constructed. In reality, the dimensionality of these points are 512 (embedding size of the ArcFace output). In this example we only show three points for clarity. This is mentioned in the main manuscript as well, but using following equation:

\begin{equation}
  z_a = \mathcal{S}_a[argmin(|cos(z_{id}, \mathcal{S}_a| + m))]~,
  \label{eq:sample}
\end{equation}

we can control a desired approximate cosine distance from the target identity by changing the margin $m$. In Figure \ref{fig:anchor} we illustrate anchor matches as the margin changes, where the green line represent the match that would occur for the margin $m$ of 0.7. Because \fii~was trained counter-factually to drive the identity away we want to find an anchor close. When using target-oriented face swapping methods, which are trained to drive the identity towards the source, you have to sample identities far away.

\begin{figure}[ht!]
\centering
\includegraphics[width=0.4\textwidth]{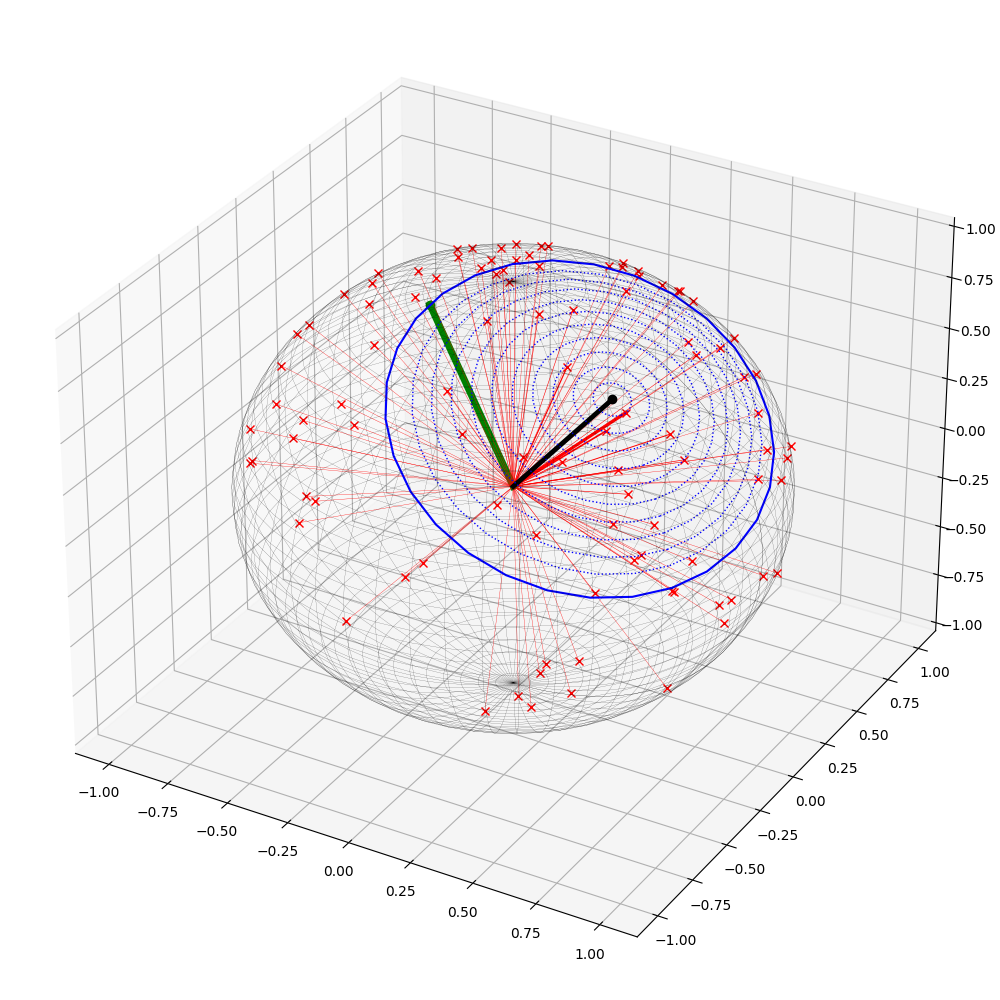}
\caption{Illustration of matching a desired anchor. The red lines illustrates matches to a desired anchor based on desired approximate distance from the target vector (black line). The green line illustrates the match that would occur for when sampling for \fii. Blue circle illustrates the desired distance.}
\label{fig:anchor}
\end{figure}

\section{Gender and Ethnicity Preservation}

Because \fii~is trained counter-factually to drive the identity away it only uses target faces, compared to face swapping which generally uses pairs of target and source faces. It makes sense that the generator learns to preserve gender and ethnicity to be able to generate convincing samples for the discriminator. We find interesting evidence for this when looking into face swapping using \fii. \fii~manage to reach state-of-the-art for identity transfer for face-swapping. However, we notice qualitatively that the gender and ethnicity it preserved, even if you would use a male target and female source or Asian target with a Caucasian source. Shown in Figure \ref{fig:attri}, we can see that \fii~tends to preserve gender and ethnicity for face swaps compared to FaceDancer. Even if \fii~performs better for identity transfer quantitatively, it does not qualitative. However this behaviour is useful for other tasks, such as anonymization, allowing us to ignore the need to sample identities based of gender.

\end{alphasection}

\end{document}